\PassOptionsToPackage{table}{xcolor}
\documentclass[sigconf]{acmart}
\usepackage{enumitem}
\usepackage{graphicx}
\AtBeginDocument{%
  }

\setcopyright{cc}
\setcctype{by}
\copyrightyear{2026}
\acmYear{2026}
\acmDOI{10.1145/3799682.3840266}
\acmConference[CIKM '26]{Proceedings of the 35th ACM International Conference on Information and Knowledge Management}{November 7--11, 2026}{Rome, Italy}
\acmBooktitle{Proceedings of the 35th ACM International Conference on Information and Knowledge Management (CIKM '26), November 7--11, 2026, Rome, Italy}
\acmISBN{979-8-4007-2539-5/2026/11}

\begin{document}

\title{PolyUQuest: Verifiable Structure-Aware Web RAG over Heterogeneous Graphs}

\settopmatter{authorsperrow=4}
\makeatletter
\def\@affiliationfont{\normalsize\normalfont}
\makeatother

\author{Ying Liu}
\affiliation{%
  \institution{The Hong Kong Polytechnic University}
  \city{Hong Kong}
  \country{China}}
\email{yarden.liu@connect.polyu.hk}

\author{Yi Ye}
\affiliation{%
  \institution{The Hong Kong Polytechnic University}
  \city{Hong Kong}
  \country{China}}
\email{yi000.ye@connect.polyu.hk}

\author{Quanyu Feng}
\affiliation{%
  \institution{The Hong Kong Polytechnic University}
  \city{Hong Kong}
  \country{China}}
\email{25061357g@connect.polyu.hk}

\author{Mingxi Ye}
\affiliation{%
  \institution{The Hong Kong Polytechnic University}
  \city{Hong Kong}
  \country{China}}
\email{mingxi.ye@connect.polyu.hk}

\author{Mingtao Zhang}
\affiliation{%
  \institution{The Hong Kong Polytechnic University}
  \city{Hong Kong}
  \country{China}}
\email{mingtao.zhang@connect.polyu.hk}

\author{Haoyang Li}
\authornote{Corresponding author.}
\affiliation{%
  \institution{The Hong Kong Polytechnic University}
  \city{Hong Kong}
  \country{China}}
\email{haoyang-comp.li@polyu.edu.hk}

\author{Chen Jason Zhang}
\affiliation{%
  \institution{The Hong Kong Polytechnic University}
  \city{Hong Kong}
  \country{China}}
\email{jason-c.zhang@polyu.edu.hk}

\author{Qing Li}
\affiliation{%
  \institution{The Hong Kong Polytechnic University}
  \city{Hong Kong}
  \country{China}}
\email{qing-prof.li@polyu.edu.hk}

\renewcommand{\shortauthors}{Liu et al.}

\begin{abstract}
Existing retrieval-augmented generation (RAG) systems treat web pages as flat text, losing the structural and semantic signals encoded in HTML. We present PolyUQuest, a verifiable, structure-aware web RAG framework built on a heterogeneous graph that unifies hyperlink topology between pages, DOM hierarchy within pages, and entity–relation knowledge across pages. A two-tier router dispatches each query to one of three retrieval modes matched to its structural need, including direct block retrieval, cross-page graph traversal, and multi-hop entity reasoning. Each answer carries traceable provenance: every cited block records its source page, heading path, and entity links, so users can inspect the structural evidence behind a claim. We evaluate on the official websites of the Hong Kong Polytechnic University (PolyU), comprising 4,240 pages, 31,086 DOM blocks, 29,119 entities, and 37,680 relations, together with a multi-type evaluation benchmark. PolyUQuest improves correctness, coverage, and faithfulness over the evaluated baselines while maintaining query-time token consumption comparable to ChunkRAG and substantially below the graph-based RAG baselines. The demonstration provides an interactive interface for inspecting cited answers, comparing retrieval traces across routing modes, and exploring evidence graph paths. PolyUQuest is being prepared for deployment as a student-facing QA service at PolyU. A demo video is available at 
\url{https://youtu.be/thKWYaL_4rw}.
\end{abstract}



\maketitle

\section{Introduction}

Web pages are among the most widely used knowledge sources for retrieval-augmented generation (RAG)~\cite{huang2025ketrag, li2025knowtrace, wang2024topologyrag}, where retrieving accurate and current evidence is key to curbing hallucination~\cite{jiang2025ras, fan2025ragdm, zhang2025siren, xu2026securingretrievalaugmentedgenerationtaxonomy}.
Unlike plain documents, web content is organized in three complementary layers: hyperlinks between pages, a DOM hierarchy within each page, and named entities that recur across pages.
Consider a prospective student asking, ``Which professors in the Department of Computing conduct NLP research and also teach related courses?''
No single page holds the answer: it is scattered across faculty profiles and course pages.
Answering it requires following hyperlinks and connecting entities across pages, beyond the reach of plain similarity retrieval.
 
Existing RAG approaches cannot handle questions over structurally complex websites, especially those that require cross-page navigation.
Plain-text and chunk-based RAG methods~\cite{singh2024chunkrag, li2025neutronrag, zhu2026skyline} flatten each page into a bag of chunks and retrieve by similarity, discarding structure entirely, while agentic web search can recover it by browsing, but only at prohibitive token cost and latency.
Graph RAG systems~\cite{wang2026agrag, guo2024lightrag, hu2025grag, liang2025kag, 10.1145/3774904.3793011}, such as LightRAG~\cite{guo2024lightrag}, instead capture entity-relation structure~\cite{peng2025graph}, yet they still flatten HTML into text and inject large global contexts that inflate the cost of every query~\cite{zhou2025graganalysis, jiang2025piperag}.
HTML- and document-structure-aware RAG~\cite{tan2025htmlrag, lin2025zendb, li2025bridgingtables}, such as HtmlRAG~\cite{tan2025htmlrag}, in contrast, preserves structure within pages or documents, but ignores the hyperlinks that connect pages and so cannot follow evidence across a site.
In summary,
existing systems fail to reason over hyperlink topology, DOM hierarchy, and cross-page entities together, and those closest to structure-awareness pay heavily in tokens and latency.


\vspace{1em}
\noindent \textbf{PolyUQuest} (Section~\ref{sec:framework}).
In response, we have developed PolyUQuest, an interactive structure-aware RAG system over heterogeneous web graphs.
PolyUQuest unifies hyperlink topology, DOM hierarchy, and cross-page entity associations under a single graph, and exploits them through query routing that addresses the tension between structural fidelity, cross-page reasoning, and retrieval costs that limit prior approaches.
Compared to existing RAG systems, PolyUQuest offers the following:

\noindent\textit{\underline{Three-layer web graph.}}
PolyUQuest models a website as a single heterogeneous graph whose webpage, evidence-block, entity, and topic nodes are connected across all three structural layers.
Unlike prior systems that usually emphasize one structural layer, this unified graph allows the system to move from an entity to the block where it appears, then to the containing page and its linked neighbors when an answer spans several pages.

\noindent\textit{\underline{Structure-driven retrieval.}}
Different questions draw on different layers of the graph, so PolyUQuest classifies each query and dispatches it to one of three retrieval modes: direct block retrieval for single-hop facts, cross-page navigation for comparison and aggregation, and multi-hop entity reasoning for evidence scattered across pages.
By retrieving only the structurally relevant evidence, this routing avoids the large global contexts that make graph RAG costly.

\noindent\textit{\underline{Verifiable answer provenance.}}
Every cited block carries its source page, heading path, and entity links.
Users can click a citation to inspect the exact page context behind a claim, follow graph paths connecting people, programmes, courses, and pages, and see at a glance whether an answer rests on one page or several.

\vspace{1em}
\noindent \textbf{Demonstration} (Section~\ref{sec:demo}).
We give a guided tour of PolyUQuest over institutional webpages, covering cited-answer inspection, guided retrieval traces, and graph-based evidence exploration.
Participants can ask questions, inspect the source blocks and pages behind each answer, compare retrieval traces across modes, and explore the graph paths that justify a result.

\section{PolyUQuest System}\label{sec:framework}

\begin{figure}[!t]
\centering
\includegraphics[width=\columnwidth]{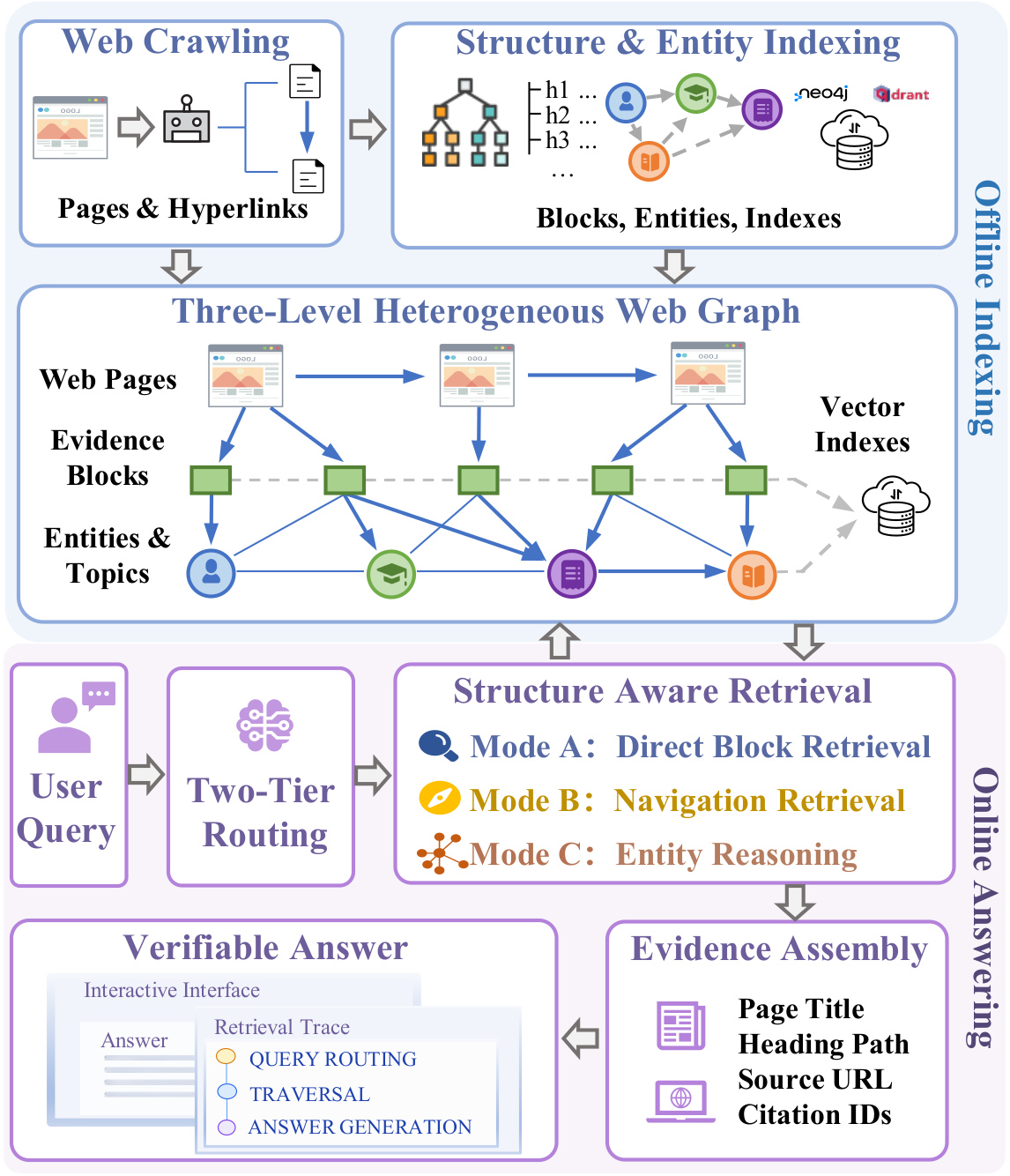}
\caption{Framework Overview of PolyUQuest}
\label{fig:architecture}
\Description{Overall architecture of PolyUQuest, including offline indexing, three-layer heterogeneous graph construction, query routing, and online retrieval.}
\end{figure}

As illustrated in Figure~\ref{fig:architecture}, this section describes offline indexing (\S\ref{sec:offline-indexing}), online retrieval (\S\ref{sec:online-retrieval}), and the full workflow (\S\ref{sec:workflow}).

\subsection{Offline Indexing}\label{sec:offline-indexing}

\noindent\textbf{Three-Layer Heterogeneous Graph.}
PolyUQuest models a website $\mathcal{W}$ as a heterogeneous graph $\mathcal{G}_{\mathcal{W}}=(\mathcal{V},\mathcal{E})$, where $\mathcal{V}=\mathcal{V}_P \cup \mathcal{V}_B \cup \mathcal{V}_E \cup \mathcal{V}_T$ contains webpage, evidence block, entity, and topic nodes~\cite{wang2024topologyrag, peng2025graph, huang2025ketrag}.
The edge set $\mathcal{E}$ captures four families of connections: page--page hyperlinks, page--block containment, block--entity mentions, and semantic associations among entities and topics.
Each evidence block node $b\in\mathcal{V}_B$ is attached to its source page $p(b)$, heading path $h(b)$, block embedding $\mathbf{x}_b$, length $\ell(b)$, and mentioned entity subset $\mathcal{V}_E(b)\subseteq\mathcal{V}_E$.
\textit{Layer~1 (Site Graph):} hyperlinks capture how webpages point to each other.
\textit{Layer~2 (Block Tree):} page content is organized into heading-aware text blocks that preserve the local document hierarchy.
\textit{Layer~3 (Entity Graph):} extracted entities are connected to their source blocks, related entities, and topics.
These layers allow the system to move from an entity to the evidence block where it appears, then to the containing page and neighboring pages when cross-page evidence is needed.

\noindent\textbf{Block Tree Construction.}
Following HTML-aware
studies
~\cite{tan2025htmlrag, lin2025zendb}, 
our system removes non-content regions such as scripts, style definitions, embedded frames, and comments from raw HTML while retaining useful textual, hyperlink, and accessibility information.
A bottom-up algorithm traverses the cleaned DOM tree, merging sibling subtrees until a block reaches a configurable word threshold, which we set to 150 words, following~\cite{tan2025htmlrag,guo2024lightrag}.
Tables, lists, and preformatted code are treated as atomic units that cannot be split.
Each retained block inherits its surrounding section headings, a compact representation that captures page hierarchy without storing the full DOM tree.
Repeated boilerplate regions, such as navigation bars, footers, and cookie notices, are filtered out before block embedding.

\noindent\textbf{Entity Extraction and Resolution.}
An LLM extracts entities and relations from each page's blocks using a small set of website-relevant entity types, such as people, programmes, departments, courses, and research topics~\cite{han2025docpolicykg, huang2024entityalignment, hashemi2025kraft}.
Each extraction is linked back to the source block, enabling fine-grained provenance.
Because the same person, programme, or department may appear under abbreviations, aliases, or slightly different page titles, PolyUQuest resolves extracted mentions before adding them to the graph.
It first normalizes obvious aliases and uses authority pages as anchors, then merges remaining candidates with a combined semantic and string similarity score; an LLM is used to decide borderline cases.

\subsection{Online Structure-Aware Retrieval}\label{sec:online-retrieval}
PolyUQuest organizes online retrieval into three modes according to the structure needed by the query: Mode~A for direct factual lookup, Mode~B for navigation across linked pages, and Mode~C for entity-based multi-hop reasoning~\cite{ye2025queryunderstanding, yu2025aquapipe}.
The router selects among these modes before the chosen retrieval pathway assembles cited evidence for answer generation.

\noindent\textbf{Two-Tier Router.}
Given a user query $q$ with embedding $\mathbf{x}_q$, the router classifies it in two tiers.
A first tier of lightweight rules handles unambiguous structural signals: ``which professors'' triggers Mode~C, ``admission requirements for'' triggers Mode~B, and comparison patterns such as ``differences between X and Y'' map to Mode~B for cross-page aggregation.
Queries that match no heuristic pass to a second-tier LLM classifier, which assigns the query to one of the three modes with a confidence score.
This design keeps routing latency low on common patterns while delegating long-tail queries to the LLM.

\noindent\textit{\underline{\textbf{Mode~A: Direct Block Retrieval.}}}
For single-hop factual questions such as ``What is the tuition fee for MSc in AI?'', the system retrieves candidates in two stages.
The first stage builds a candidate block set $\mathcal{B}_q$ by combining dense approximate nearest-neighbor search over $\mathbf{x}_b$ with BM25 sparse retrieval, capturing both semantic and lexical signals~\cite{li2025neutronrag, yu2025pruningrag}.
This mode treats each heading-aware block as the basic evidence unit, so it is suitable when the answer is likely localized within one page section.
A cross-encoder reranker then rescores blocks in $\mathcal{B}_q$ by jointly attending to the query, the block text, its source page $p(b)$, and heading path $h(b)$.
The top blocks are enriched with parent block content before answer generation, which helps preserve local context for citations.

\noindent\textit{\underline{\textbf{Mode~B: Navigation Retrieval.}}}
For aggregation queries across pages, such as comparing two programmes, the system decomposes the query into subqueries via the LLM.
For each subquery, it first retrieves relevant pages and then expands to nearby linked pages in the site graph.
The candidate set $\mathcal{B}_q$ is then assembled from evidence blocks on the discovered pages.
This expansion is important for university websites and other structured web collections, where overview pages often link to separate pages for fees, admission, deadlines, or contacts~\cite{chen2024msmarcoweb, wang2024topologyrag, choi2025referencealigned}.
A holistic cross-encoder reranking scores all candidates across subqueries before the final synthesis, allowing the answer generator to compare evidence from multiple pages under a shared context.

\noindent\textit{\underline{\textbf{Mode~C: Entity Reasoning.}}}
For queries requiring multiple hops, such as ``Which professors research NLP and teach related courses?'', the system first extracts topic keywords via the LLM and then follows two complementary evidence paths.
The first path retrieves candidate entities and expands through entity relations.
The second path retrieves topic keywords and expands to entities associated with those topics.
Together, these paths identify query-related entity nodes $\mathcal{V}_{E,q}\subseteq\mathcal{V}_E$, and the system traces them back to source blocks that mention them~\cite{quan2025thoughtforest, li2025knowtrace, peng2025graph, li2025kug}.
This mode is designed for questions whose evidence is scattered across multiple pages rather than concentrated in a single document.
Using the candidate blocks $\mathcal{B}_q$, Mode~C ranks each block $b$ by:
\begin{equation}
\begin{aligned}
s(b,q) ={}&  \cos(\mathbf{x}_q,\mathbf{x}_b)
+  \frac{\min(|\mathcal{V}_E(b) \cap \mathcal{V}_{E,q}|,\kappa)}{\kappa} \\
&+  \left(1 - \frac{\ell(b)}{\max_{b' \in \mathcal{B}_q} \ell(b')}\right).
\end{aligned}
\end{equation}
\noindent 
The three terms measure query--block similarity, capped entity coverage, and block conciseness.
The cap $\kappa{=}30$, tuned on validation sets, prevents blocks with many incidental entities from dominating the ranking.
After scoring, a same-page penalty encourages source diversity when many retrieved blocks come from the same page.

\vspace{-0.5em}

\subsection{Workflow}\label{sec:workflow}

PolyUQuest has two phases: an offline phase that builds a three-layer graph from crawled pages, and an online phase that routes queries to structure-aware retrieval pathways. 

\noindent\textbf{Offline Indexing Pipeline.}

\begin{itemize}[leftmargin=*, itemsep=0pt, parsep=0pt, topsep=1pt]
\item \textit{\underline{Web Crawling.}}
PolyUQuest crawls selected PolyU webpages and records hyperlinks between pages as the site graph.
\item \textit{\underline{DOM Evidence Construction.}}
It parses each page into heading-aware evidence blocks and preserves section context for later citation and evidence inspection.
\item \textit{\underline{Entity Graph Construction.}}
It extracts entities and relations from evidence blocks, resolves duplicate entities, and links each entity back to its source evidence.
\end{itemize}

\noindent\textbf{Online Query Pipeline.}
\begin{itemize}[leftmargin=*, itemsep=0pt, parsep=0pt, topsep=1pt]
\item \textit{\underline{Query Routing.}}
The router assigns the query to factual lookup, navigation across pages, or entity reasoning.
\item \textit{\underline{Structure-Aware Retrieval.}}
The selected mode retrieves evidence from DOM blocks, linked pages, or entity paths.
\item \textit{\underline{Evidence Assembly.}}
The system reranks retrieved blocks and assembles cited evidence with page titles and heading paths.
\item \textit{\underline{Answer Generation.}}
The LLM generates an answer while the interface displays citations and the retrieval trace.
\end{itemize}

\begin{figure*}[!t]
\centering
 \includegraphics[width=\textwidth, trim=0cm 0.2cm 0cm 0.2cm, clip]{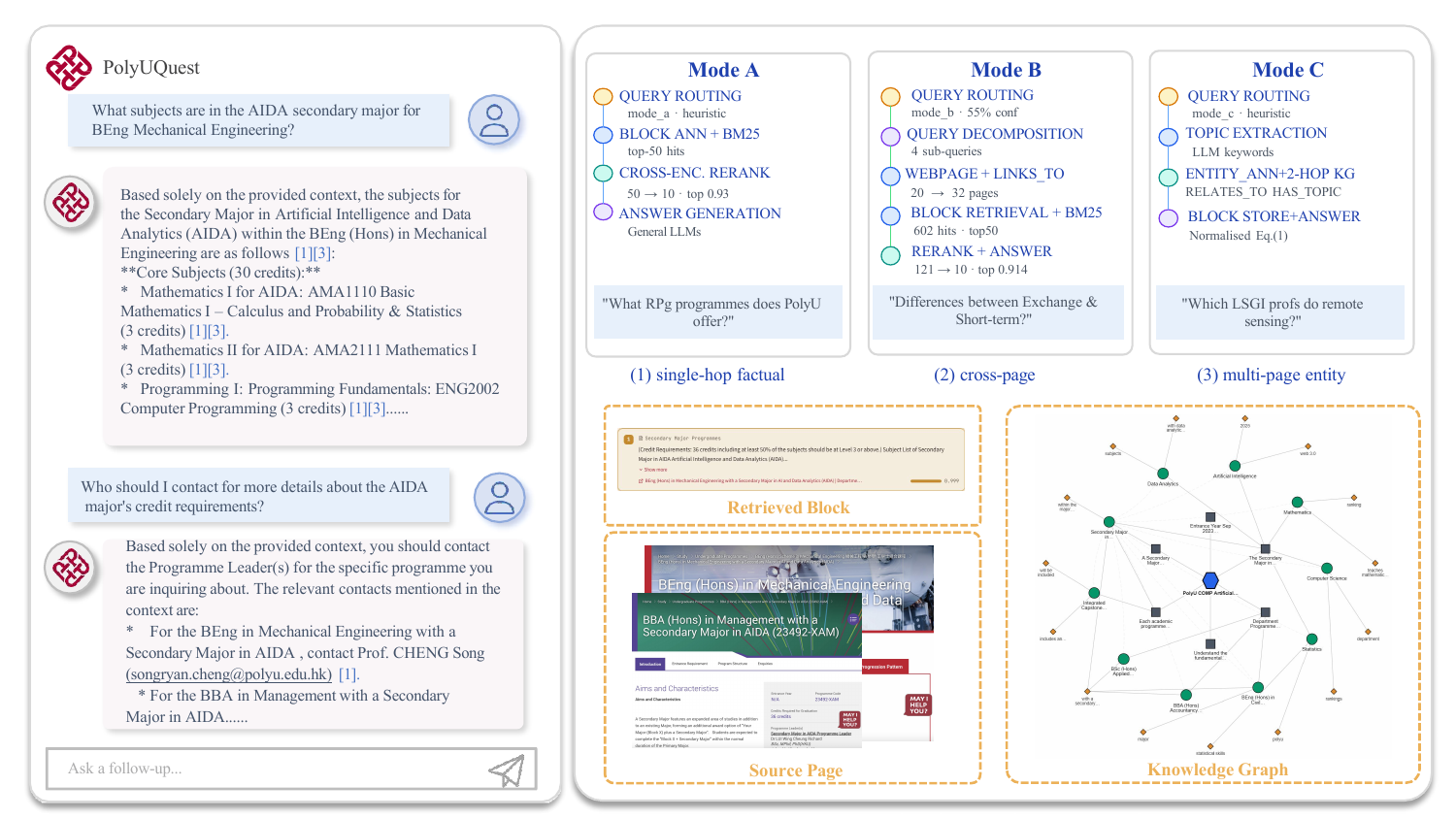}
\caption{Interactive demo interface of PolyUQuest.}
\label{fig:process}
\Description{PolyUQuest demo interface showing the chat view, retrieval traces for the three routing modes, source-page evidence, and graph-based provenance paths.}
\end{figure*}

\section{Demonstration}\label{sec:demo}

\noindent\textbf{Demo Setup.}
The demo runs on 4,240 official English pages from the PolyU website, covering programme, staff, department, and research sections, with 31,086 DOM blocks, 29,119 resolved entities, and 37,680 relations. The system provides a Chat view for arbitrary question answering with cited evidence and a Graph view for exploring the underlying webpage, DOM, and entity structures. The source code is in \url{https://github.com/13-pieces-teen/PolyUQuest}.


\smallskip
\noindent\textbf{Demonstration Scenarios.}
We will guide users through three scenarios: answer provenance inspection, guided retrieval trace, and evidence graph exploration. Figure~\ref{fig:process} illustrates the three demonstration scenarios:
the Chat view with cited answers and follow-up questions (left), the retrieval traces for each routing mode (top), and the evidence inspection panels showing source pages, retrieved blocks, and knowledge graph paths (bottom right).

\noindent\textit{\underline{Scenario~1: Answer Provenance Inspection.}}
This scenario covers the primary question answering use case.
Users first enter a programme or staff-related question in the Chat view.
PolyUQuest returns a concise answer with inline citations, while the evidence panel shows the source page, heading path, and DOM block behind each cited claim.
Users can click a citation to highlight the corresponding evidence block and inspect its surrounding page context.
The interface also groups evidence by source page, so users can quickly see whether an answer is grounded in one page or assembled from several pages.
This scenario shows that users can verify the answer without manually opening and comparing many webpages.

\noindent\textit{\underline{Scenario~2: Guided Retrieval Trace.}}
Users then try three representative queries: a single-page factual question, a cross-page comparison, and an entity-centric question.
For each query, the trace panel shows the selected retrieval mode and summarizes the main steps from routing to evidence assembly.
Users can compare how the trace changes across factual lookup, page navigation, and entity reasoning.
The trace also reports the retrieved pages and evidence blocks at each stage, making routing decisions and retrieval failures visible.
This scenario demonstrates how different user intents lead to different retrieval behaviors while keeping the interface understandable to non-expert users.

\noindent\textit{\underline{Scenario~3: Evidence Graph Exploration.}}
Finally, users open the Graph view from a cited answer.
The interface highlights the webpage, DOM block, entity, and topic nodes that support the current answer.
Users can expand neighboring nodes to explore related pages or entities.
Hovering over nodes reveals page titles, block snippets, or entity descriptions, while selected paths remain highlighted against the cited answer.
This scenario shows how PolyUQuest turns an answer into an inspectable evidence path.

\smallskip

\noindent\textbf{Performance Highlights.}
We evaluate PolyUQuest on 300 annotated PolyU-Web questions spanning factual, cross-page, and entity-centric retrieval. Four RAG baselines use the same generator, prompt, and embedding model. We report correctness, coverage, faithfulness, and query/build token costs. The upper block of Table~\ref{tab:perf} shows that PolyUQuest leads all quality metrics at low token cost through targeted routing.
The lower block shows that DOM-aware blocks contribute most to correctness and coverage, while cross-page navigation provides smaller but consistent gains. Faithfulness remains high across both variants, indicating that structure-aware segmentation is the main source of PolyUQuest's quality advantage.

\begin{table}[t!]
\centering\small
\caption{Main results and ablation on PolyU-Web (300 questions). Q.\,Tok.: average LLM tokens per query; B.\,Tok.: total offline build tokens; ``--'': no offline build phase.}
\label{tab:perf}
\resizebox{\columnwidth}{!}{%
\begin{tabular}{lccc|cc}
\toprule
\textbf{System} & \textbf{Corr.} & \textbf{Cov.} & \textbf{Faith.} & \textbf{Q. Tok.\,$\downarrow$} & \textbf{B. Tok.\,$\downarrow$} \\
\midrule
ChunkRAG~\cite{singh2024chunkrag}           & 0.532 & 0.479 & 0.710 & 2,947 & --     \\
HtmlRAG~\cite{tan2025htmlrag}   & 0.453 & 0.448 & 0.804 & 4,009      & --     \\
FastGraphRAG~\cite{circlemind2024fastgraphrag} & 0.295 & 0.469 & 0.737 & 4,484      & 28.1M     \\
LightRAG~\cite{guo2024lightrag}  & 0.610 & 0.612 & 0.559 & 29,825 & 37.4M \\
\midrule
\rowcolor{gray!20}
\textbf{PolyUQuest}     & \textbf{0.644} & \textbf{0.649} & \textbf{0.921} & 2,968 & \textbf{17.5M} \\
\quad w/o DOM blocks         & 0.563 & 0.510 & 0.916 & 2,657 & \textbf{17.5M} \\
\quad w/o cross-page mode    & 0.624 & 0.633 & 0.905 & \textbf{2,130} & \textbf{17.5M} \\
\bottomrule
\end{tabular}%
}
 \end{table}

\smallskip
\noindent\textbf{Generalizability Discussion.} PolyUQuest targets organizational websites whose knowledge spans linked pages, hierarchical sections, and recurring entities, including universities, government agencies, hospitals, and enterprises. Adapting it to a new domain requires a fresh crawl and entity schema, while the indexing, routing, retrieval, and provenance pipeline remains unchanged.

\section{Conclusion}
We present PolyUQuest, a verifiable structure-aware web RAG system that unifies hyperlink topology, DOM hierarchy, and cross-page entity relations in one heterogeneous graph. Structure-matched routing retrieves targeted evidence for factual, cross-page, and entity-centric queries with traceable provenance. The demo lets attendees ask questions, inspect the blocks and graph traces behind each answer, and compare retrieval modes. PolyUQuest is being prepared for deployment as a QA service at PolyU.

\vspace{-0.5em}
\begin{acks}
This work was partially supported by the Guangdong Basic and Applied Basic Research Foundation (2026A1515010227) and PolyU Teaching Development Grants (TDG25-28/TBP/11-IICA).
\end{acks}

\section*{GenAI Usage Disclosure}
Generative AI tools (Anthropic Claude) were used solely to improve the grammar, spelling, and clarity of author-written text. They were not used to generate any substantive content, and the authors take full responsibility for the entire work.

\bibliographystyle{ACM-Reference-Format}
\balance
\bibliography{reference}

@inproceedings{jiang2025ras,
  title={Retrieval and structuring augmented generation with large language models},
  author={Jiang, Pengcheng and Ouyang, Siru and Jiao, Yizhu and Zhong, Ming and Tian, Runchu and Han, Jiawei},
  booktitle={Proceedings of the 31st ACM SIGKDD Conference on Knowledge Discovery and Data Mining V. 2},
  pages={6032--6042},
  year={2025}
}

@inproceedings{huang2025ketrag,
  title={Ket-rag: A cost-efficient multi-granular indexing framework for graph-rag},
  author={Huang, Yiqian and Zhang, Shiqi and Xiao, Xiaokui},
  booktitle={Proceedings of the 31st ACM SIGKDD Conference on Knowledge Discovery and Data Mining V. 2},
  pages={1003--1012},
  year={2025}
}

@inproceedings{li2025knowtrace,
  title={Knowtrace: Bootstrapping iterative retrieval-augmented generation with structured knowledge tracing},
  author={Li, Rui and Dai, Quanyu and Zhang, Zeyu and Chen, Xu and Dong, Zhenhua and Wen, Ji-Rong},
  booktitle={Proceedings of the 31st ACM SIGKDD Conference on Knowledge Discovery and Data Mining V. 2},
  pages={1470--1480},
  year={2025}
}

@inproceedings{jiang2025piperag,
  title={Piperag: Fast retrieval-augmented generation via adaptive pipeline parallelism},
  author={Jiang, Wenqi and Zhang, Shuai and Han, Boran and Wang, Jie and Wang, Bernie and Kraska, Tim},
  booktitle={Proceedings of the 31st ACM SIGKDD Conference on Knowledge Discovery and Data Mining V. 1},
  pages={589--600},
  year={2025}
}

@article{yu2025aquapipe,
  title={AquaPipe: A Quality-Aware Pipeline for Knowledge Retrieval and Large Language Models},
  author={Yu, Runjie and Huang, Weizhou and Bai, Shuhan and Zhou, Jian and Wu, Fei},
  journal={Proceedings of the ACM on Management of Data},
  volume={3},
  number={1},
  pages={1--26},
  year={2025},
  publisher={ACM New York, NY, USA}
}

@inproceedings{li2025neutronrag,
  title={NeutronRAG: Towards Understanding the Effectiveness of RAG from a Data Retrieval Perspective},
  author={Li, Peizheng and Chen, Chaoyi and Yuan, Hao and Fu, Zhenbo and Shen, Hang and Yang, Xinbo and Wang, Qiange and Ai, Xin and Zhang, Yanfeng and Wen, Yingyou and others},
  booktitle={Companion of the 2025 International Conference on Management of Data},
  pages={163--166},
  year={2025}
}

@inproceedings{fan2025ragdm,
  title={Towards Retrieval-Augmented Large Language Models: Data Management and System Design},
  author={Fan, Wenqi and Wu, Pangjing and Ding, Yujuan and Ning, Liangbo and Wang, Shijie and Li, Qing},
  booktitle={2025 IEEE 41st International Conference on Data Engineering (ICDE)},
  pages={4509--4512},
  year={2025},
  organization={IEEE}
}

@inproceedings{lin2025zendb,
  title={Querying templatized document collections with large language models},
  author={Lin, Yiming and Hulsebos, Madelon and Ma, Ruiying and Shankar, Shreya and Zeighami, Sepanta and Parameswaran, Aditya G and Wu, Eugene},
  booktitle={2025 IEEE 41st International Conference on Data Engineering (ICDE)},
  pages={2422--2435},
  year={2025},
  organization={IEEE}
}

@inproceedings{huang2024entityalignment,
  title={Representation learning for entity alignment in knowledge graph: A design space exploration},
  author={Huang, Peng and Zhang, Meihui and Zhong, Ziyue and Chai, Chengliang and Fan, Ju},
  booktitle={2024 IEEE 40th International Conference on Data Engineering (ICDE)},
  pages={3462--3475},
  year={2024},
  organization={IEEE}
}

@inproceedings{li2025kug,
  title={KUG: Joint Enhancement of Internal and External Knowledge for Retrieval-Augmented Generation},
  author={Li, Mingyang and Chen, Shisong and Tu, Shengkun and Du, Ziyi and Zhang, Jinghao and Li, Zhixu and Xiao, Yanghua},
  booktitle={Proceedings of the 34th ACM International Conference on Information and Knowledge Management},
  pages={1603--1612},
  year={2025}
}

@inproceedings{choi2025referencealigned,
  title={Reference-Aligned Retrieval-Augmented Question Answering over Heterogeneous Proprietary Documents},
  author={Choi, Nayoung and Byun, Grace and Chung, Andrew and Paek, Ellie S and Lee, Shinsun and Choi, Jinho D},
  booktitle={Proceedings of the 34th ACM International Conference on Information and Knowledge Management},
  pages={5626--5633},
  year={2025}
}

@inproceedings{yu2025pruningrag,
  title={Multi-source knowledge pruning for retrieval-augmented generation: A benchmark and empirical study},
  author={Yu, Shuo and Cheng, Mingyue and Liu, Qi and Wang, Daoyu and Yang, Jiqian and Ouyang, Jie and Luo, Yucong and Lei, Chenyi and Chen, Enhong},
  booktitle={Proceedings of the 34th ACM International Conference on Information and Knowledge Management},
  pages={3931--3941},
  year={2025}
}

@inproceedings{ye2025queryunderstanding,
  title={Can LLMs Really Help Query Understanding In Web Search? A Practical Perspective},
  author={Ye, Dezhi and Qin, Ye and Tian, Bowen and Fan, Jiabin and Liu, Jie and Liang, Haijin and Ma, Jin},
  booktitle={Proceedings of the 34th ACM International Conference on Information and Knowledge Management},
  pages={5433--5438},
  year={2025}
}

@inproceedings{li2025bridgingtables,
  title={Bridging Queries and Tables through Entities in Open-Domain Table Retrieval},
  author={Li, Da and Bi, Keping and Guo, Jiafeng and Cheng, Xueqi},
  booktitle={Proceedings of the 34th ACM International Conference on Information and Knowledge Management},
  pages={1540--1550},
  year={2025}
}

@inproceedings{han2025docpolicykg,
  title={DocPolicyKG: A Lightweight LLM-Based Framework for Knowledge Graph Construction from Chinese Policy Documents},
  author={Han, Chen and Li, Yuanyuan and Tang, Xijin},
  booktitle={Proceedings of the 34th ACM International Conference on Information and Knowledge Management},
  pages={4753--4757},
  year={2025}
}

@inproceedings{quan2025thoughtforest,
  title={ThoughtForest-KGQA: A Multi-Chain Tree Search for Knowledge Graph Reasoning},
  author={Quan, Xingrun and Zhou, Yongkang and Yao, Junjie},
  booktitle={Proceedings of the 34th ACM International Conference on Information and Knowledge Management},
  pages={5156--5160},
  year={2025}
}

@inproceedings{hashemi2025kraft,
  title={KRAFT: A Knowledge Graph-Based Framework for Automated Map Conflation},
  author={Hashemi, Farnoosh and Lakshmanan, Laks VS},
  booktitle={Proceedings of the 34th ACM International Conference on Information and Knowledge Management},
  pages={802--812},
  year={2025}
}

@inproceedings{wang2024topologyrag,
  title={Topology-aware retrieval augmentation for text generation},
  author={Wang, Yu and Lipka, Nedim and Zhang, Ruiyi and Siu, Alexa and Zhao, Yuying and Ni, Bo and Wang, Xin and Rossi, Ryan and Derr, Tyler},
  booktitle={Proceedings of the 33rd ACM International Conference on Information and Knowledge Management},
  pages={2442--2452},
  year={2024}
}

@inproceedings{chen2024msmarcoweb,
  title={Ms marco web search: A large-scale information-rich web dataset with millions of real click labels},
  author={Chen, Qi and Geng, Xiubo and Rosset, Corby and Buractaon, Carolyn and Lu, Jingwen and Shen, Tao and Zhou, Kun and Xiong, Chenyan and Gong, Yeyun and Bennett, Paul and others},
  booktitle={Companion Proceedings of the ACM Web Conference 2024},
  pages={292--301},
  year={2024}
}

@inproceedings{tan2025htmlrag,
  title={Htmlrag: Html is better than plain text for modeling retrieved knowledge in rag systems},
  author={Tan, Jiejun and Dou, Zhicheng and Wang, Wen and Wang, Mang and Chen, Weipeng and Wen, Ji-Rong},
  booktitle={Proceedings of the ACM on Web Conference 2025},
  pages={1733--1746},
  year={2025}
}

@article{guo2024lightrag,
  title={Lightrag: Simple and fast retrieval-augmented generation},
  author={Guo, Zirui and Xia, Lianghao and Yu, Yanhua and Ao, Tian and Huang, Chao},
  journal={arXiv preprint arXiv:2410.05779},
  volume={2},
  number={3},
  year={2024}
}

@inproceedings{hu2025grag,
   title={Grag: Graph retrieval-augmented generation},
  author={Hu, Yuntong and Lei, Zhihan and Zhang, Zheng and Pan, Bo and Ling, Chen and Zhao, Liang},
  booktitle={Findings of the Association for Computational Linguistics: NAACL 2025},
  pages={4145--4157},
  year={2025}
}

@inproceedings{liang2025kag,
   title={Kag: Boosting llms in professional domains via knowledge augmented generation},
  author={Liang, Lei and Bo, Zhongpu and Gui, Zhengke and Zhu, Zhongshu and Zhong, Ling and Zhao, Peilong and Sun, Mengshu and Zhang, Zhiqiang and Zhou, Jun and Chen, Wenguang and others},
  booktitle={Companion Proceedings of the ACM on Web Conference 2025},
  pages={334--343},
  year={2025}
}

@article{peng2025graph,
  title={Graph retrieval-augmented generation: A survey},
  author={Peng, Boci and Zhu, Yun and Liu, Yongchao and Bo, Xiaohe and Shi, Haizhou and Hong, Chuntao and Zhang, Yan and Tang, Siliang},
  journal={ACM Transactions on Information Systems},
  volume={44},
  number={2},
  pages={1--52},
  year={2025},
  publisher={ACM New York, NY}
}

@article{zhang2025siren,
  title={Siren’s Song in the AI Ocean: A Survey on Hallucination in Large Language Models},
  author={Zhang, Yue and Li, Yafu and Cui, Leyang and Cai, Deng and Liu, Lemao and Fu, Tingchen and Huang, Xinting and Zhao, Enbo and Zhang, Yu and Chen, Yulong and others},
  journal={Computational Linguistics},
  volume={51},
  number={4},
  pages={1373--1418},
  year={2025},
  publisher={MIT Press 255 Main Street, 9th Floor, Cambridge, Massachusetts 02142, USA~…}
}

@article{zhou2025graganalysis,
  title={In-Depth Analysis of Graph-Based RAG in a Unified Framework},
  author={Zhou, Yingli and Su, Yaodong and Sun, Youran and Wang, Shu and Wang, Taotao and He, Runyuan and Zhang, Yongwei and Liang, Sicong and Liu, Xilin and Ma, Yuchi and others},
  journal={Proceedings of the VLDB Endowment},
  volume={18},
  number={13},
  pages={5623--5637},
  year={2025},
  publisher={VLDB Endowment}
}

@article{singh2024chunkrag,
  title={Chunkrag: Novel llm-chunk filtering method for rag systems},
  author={Singh, Ishneet Sukhvinder and Aggarwal, Ritvik and Allahverdiyev, Ibrahim and Taha, Muhammad and Akalin, Aslihan and Zhu, Kevin and O'Brien, Sean},
  journal={arXiv preprint arXiv:2410.19572},
  year={2024}
}

@misc{circlemind2024fastgraphrag,
  author = {Circlemind},
  title = {Fast {GraphRAG}},
  year = {2024},
  url = {https://github.com/circlemind-ai/fast-graphrag},
}

@misc{xu2026securingretrievalaugmentedgenerationtaxonomy,
      title={Securing Retrieval-Augmented Generation: A Taxonomy of Attacks, Defenses, and Future Directions}, 
      author={Yuming Xu and Mingtao Zhang and Zhuohan Ge and Haoyang Li and Nicole Hu and Yongqi Zhang and Zhiyuan Wen and Jason Chen Zhang and Qing Li and Lei Chen},
      year={2026},
      eprint={2604.08304},
      archivePrefix={arXiv},
      primaryClass={cs.CR},
      url={https://arxiv.org/abs/2604.08304}, 
}

@inproceedings{10.1145/3774904.3793011,
author = {Wang, Yubo and Li, Haoyang and Teng, Fei and Chen, Lei},
title = {GORAG: Graph-based Online Retrieval Augmented Generation for Dynamic Few-shot Social Media Text Classification},
year = {2026},
isbn = {9798400723070},
publisher = {Association for Computing Machinery},
address = {New York, NY, USA},
url = {https://doi.org/10.1145/3774904.3793011},
doi = {10.1145/3774904.3793011},
booktitle = {Proceedings of the ACM Web Conference 2026},
pages = {9148–9159},
numpages = {12},
location = {United Arab Emirates},
series = {WWW '26}
}

@article{zhu2026skyline,
  title={Skyline Retrieval meets Set-Cover Chunk Merging: A Cost-Effective {RAG}-Sketch for Long-Context {LLM} {QA}},
  author={Zhu, Xinyi and Li, Haoyang and Zhang, Yongqi and Chen, Lei},
  journal={Proceedings of the ACM on Management of Data},
  volume={4},
  number={3 (SIGMOD)},
  pages={1--27},
  year={2026},
  publisher={ACM New York, NY, USA}
}

@inproceedings{wang2026agrag,
  title={{AGRAG}: Advanced graph-based retrieval-augmented generation for {LLMs}},
  author={Wang, Yubo and Li, Haoyang and Teng, Fei and Chen, Lei},
  booktitle={2026 IEEE 42nd International Conference on Data Engineering (ICDE)},
  pages={1--15},
  year={2026},
  organization={IEEE}
}

\end{document}